\documentclass[10pt,twocolumn,letterpaper]{article}

\usepackage{wacv}              

\usepackage{graphicx}
\usepackage{amsmath}
\usepackage{amssymb}
\usepackage{booktabs}
\usepackage{wacv}
\usepackage{times}
\usepackage{epsfig}

\usepackage{xcolor}
\usepackage{subcaption}

\usepackage{bbm}
\usepackage{mathrsfs}
\usepackage{color, colortbl}

\definecolor{Gray}{gray}{0.9}
\definecolor{LightCyan}{rgb}{0.88,1,1}
\newcolumntype{a}{>{\columncolor{LightCyan}}c}

\usepackage[pagebackref,breaklinks,colorlinks]{hyperref}

\usepackage[capitalize]{cleveref}
\crefname{section}{Section}{Secs.}
\Crefname{section}{Section}{Sections}
\Crefname{table}{Table}{Tables}
\crefname{table}{Tab.}{Tabs.}

\def\ie{\emph{i.e.}}
\def\eg{\emph{e.g.}}

\def\etal{\emph{et al.~}}

\def\wacvPaperID{42} 
\def\confName{WACV}
\def\confYear{2024}

\begin{document}

\title{LibreFace: An Open-Source  Toolkit for Deep Facial Expression Analysis}


\author{Di Chang, 
Yufeng Yin,
Zongjian Li,  
Minh Tran,
Mohammad Soleymani
\\
Institute for Creative Technologies, University of Southern California \\
{\tt\small \{dichang,yufengy,minhntra\}@usc.edu, \{lizongjian,soleymani\}@ict.usc.edu }
}

\maketitle
\thispagestyle{empty}

\begin{abstract}

Facial expression analysis is an important tool for human-computer interaction. In this paper, we introduce LibreFace, an open-source toolkit for facial expression analysis. This open-source toolbox offers real-time and offline analysis of facial behavior through deep learning models, including facial action unit (AU) detection, AU intensity estimation, and facial expression recognition. To accomplish this, we employ several techniques, including the utilization of a large-scale pre-trained network, feature-wise knowledge distillation, and task-specific fine-tuning. These approaches are designed to effectively and accurately analyze facial expressions by leveraging visual information, thereby facilitating the implementation of real-time interactive applications. In terms of Action Unit (AU) intensity estimation, we achieve a Pearson Correlation Coefficient (PCC) of \textbf{\textit{0.63}} on DISFA, which is \textbf{\textit{7\%}} higher than the performance of OpenFace 2.0~\cite{baltrusaitis2018openface} while maintaining highly-efficient inference that runs \textbf{two times} faster than OpenFace 2.0~\cite{baltrusaitis2018openface}. Despite being compact, our model also demonstrates competitive performance to state-of-the-art facial expression analysis methods on AffecNet, FFHQ, and RAF-DB. Our code will be released at \url{https://github.com/ihp-lab/LibreFace}
\end{abstract}

\begin{figure}[t]
\centering
  \includegraphics[width=0.9\linewidth]{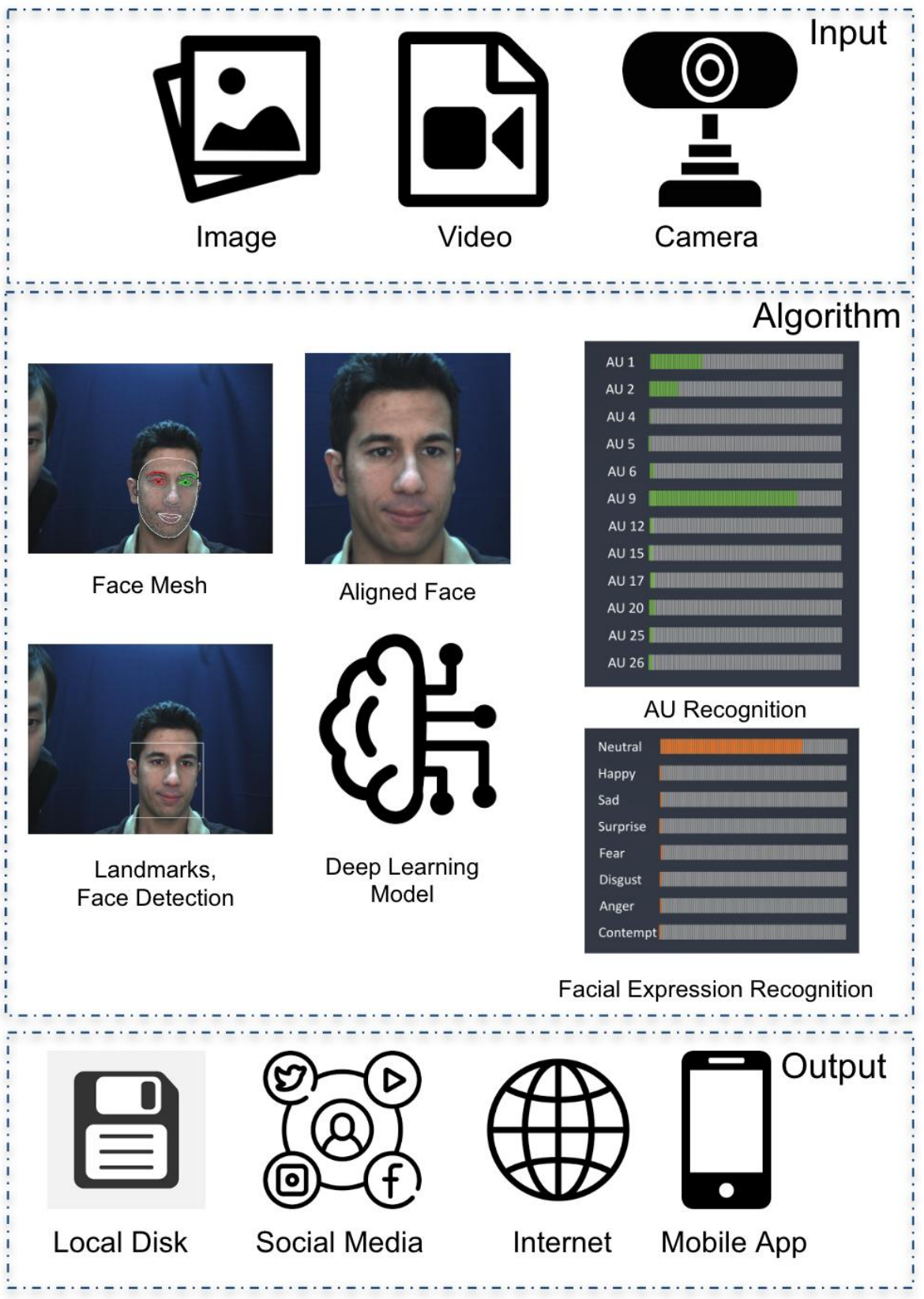}
  \caption{LibreFace is an open-source, accurate, and efficient framework for facial expression analysis, including: face mesh detection, landmark detection, AU intensity estimation, and facial expression recognition. LibreFace supports CPU-only environment as well as GPU acceleration when necessary.}
  \label{fig:system}
  \vspace{-20pt}
\end{figure}

\section{Introduction}
Facial expression analysis is the process of automatic detection of subtle facial muscle movements and recognition of prototypical facial displays. Recognizing facial expressions can provide valuable information about users' social and affective states, with a wide range of applications in human-computer interaction (HCI) to mental health \cite{gratch2014distress, devault2014simsensei}, or human-agent negotiation \cite{de2014humans}.

Two widely used methods for facial expression analysis are facial Action Unit (AU) intensity estimation \cite{ekman1977facial} and facial expression recognition. A facial action unit is an indicator of activation of an individual or a group of muscles, \eg, cheek raiser (AU6). AUs are formalized by Paul Ekman in Facial Action Coding System (FACS) \cite{ekman1977facial}. Accurate and efficient AU intensity estimation is crucial for recognizing complex facial expressions. Furthermore, facial expression recognition (FER) refers to identifying the prototypical facial displays or expressions, \eg, expressions of sadness. This is a challenging problem due to the high complexity and variability of facial expressions.

\begin{figure}[t]
\centering
  \includegraphics[width=0.9\linewidth]{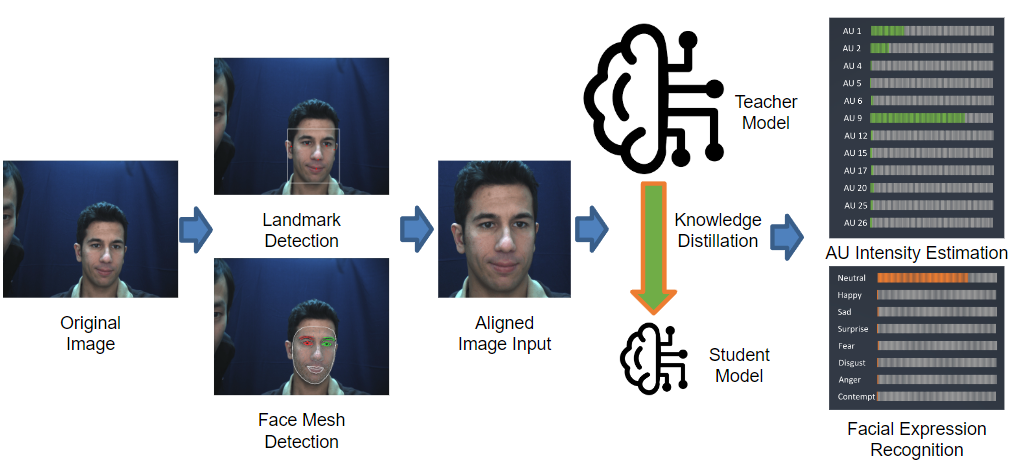}
  \caption{Overall pipeline of our proposed LibreFace, including face mesh and landmark detection, image alignment, action unit recognition, and facial expression recognition. All outputs from the pipeline can be saved, and training/inference code is open-sourced.}
  \label{fig:pipeline}
  \vspace{-20pt}
\end{figure}

Over the past few years, previous studies on facial expression analysis have achieved promising performance with deep learning-based methods \cite{shao2019facial, shao2021jaa, kollias2021distribution}. However, recent facial expression tools (see Table~\ref{tab:toolboxes}), \eg OpenFace \cite{baltruvsaitis2016openface} and OpenFace 2.0 \cite{baltrusaitis2018openface} are developed with traditional machine learning methods such as Support Vector Machine (SVM), which are less accurate than the recent deep models. 
On the other hand, despite achieving superior performance, state-of-the-art deep learning-based models are large networks with time-consuming training and inference requirements, which significantly prevent the deployment of such methods in real-world scenarios.

To address the aforementioned problems, we introduce LibreFace, an open-source, accurate, and efficient framework for facial expression analysis (see Figure~\ref{fig:system}). Specifically, we develop AU intensity estimation and facial expression recognition models with the SOTA pre-trained vision encoders, \eg, ResNet \cite{he2016deep}, Swin-Transformer \cite{xu2021generative}, and Masked Auto-encoder (MAE) \cite{he2022masked}. To boost the performance, we further pre-train the vision encoders on various face datasets, including EmotioNet \cite{fabian2016emotionet}, AffectNet \cite{kollias2021affect}, and FFHQ \cite{karras2019style}. Furthermore, to achieve real-time facial expression analysis, we utilize the feature-wise knowledge distillation based on \cite{zhang2021improve, yang2021knowledge} to improve the inference efficiency. We deliver an executable cross-platform software that supports both CPU-only and GPU-accelerated solutions according to users' hardware.

Extensive experiments show that LibreFace provides a more accurate, comprehensive, and efficient alternative to OpenFace 2.0~\cite{baltrusaitis2018openface}, the most commonly used facial expression analysis toolkit, and other open facial behavior analysis software. For AU intensity estimation, LibreFace achieves superior performance than OpenFace 2.0~\cite{baltrusaitis2018openface} while running two times faster on a CPU-only environment. For FER, LibreFace can achieve competitive performance to the heavy state-of-the-art methods.

Our major contributions are summarized as follows.
\begin{itemize}
\item We present LibreFace, an open-source and comprehensive toolkit for accurate and real-time facial expression analysis with both CPU-only and GPU-acceleration versions. LibreFace eliminates the gap between cutting-edge research and an easy and free-to-use non-commercial toolbox.

\item We propose to adaptively pre-train the vision encoders with various face datasets and then distillate them to a lightweight ResNet-18 model in a feature-wise matching manner.

\item We conduct extensive experiments of pre-training and distillation to demonstrate that our proposed pipeline achieves comparable results to state-of-the-art works while maintaining real-time efficiency.

\item LibreFace system supports cross-platform running, and the code is open-sourced in C\# (model inference and checkpoints) and Python (model training, inference, and checkpoints).

\end{itemize}

\begin{table*}[t]
    \small
    \centering
    \caption{Comparison of facial behavior analysis tools. \textit{AU} and \textit{FER} stand for AU intensity estimation and facial expression recognition. Note that the checkmark for \textit{Test} requires releasing of both checkpoints and codes for inference. \textit{Executable} denotes if the corresponding toolbox releases a binary executable program. *: Toolbox without \checkmark does not support GPU for acceleration purposes, but can still be running on CPU-only environment.}
    \resizebox{\linewidth}{!}{\begin{tabular}{l|lcccccccccccc}
    \toprule
    \rowcolor{Gray}  
    Toolbox & Approach & Landmark & GPU support* & AU & FER & Train & Test & Executable & Real-time & Free \\
    \midrule
    dlib~\cite{dlib09} & ~\cite{kazemi2014one} & \checkmark &  & & & \checkmark & \checkmark & & \checkmark & \checkmark\\
    \midrule
    FaceTracker & ~\cite{saragih2011deformable} & \checkmark &  & & & & \checkmark & \checkmark & \checkmark & \checkmark\\
    \midrule
    Mediapipe~\cite{48292} & ~\cite{kartynnik2019real} & \checkmark & \checkmark & &  \checkmark&  & \checkmark & \checkmark & \checkmark & \checkmark \\
    \midrule
    AFFDEX & unknown & \checkmark &  & \checkmark & \checkmark & & & \checkmark & \checkmark & \\
    \midrule
    AFFDEX 2.0~\cite{bishay2022affdex} & CNN,~\cite{ren2015faster} & \checkmark &  & \checkmark & \checkmark & & & \checkmark & \checkmark & \\
    \midrule
    FACET & unknown& \checkmark &  & \checkmark & & & & \checkmark & \checkmark & \\
    \midrule
    OpenFace~\cite{baltruvsaitis2016openface} & ~\cite{baltruvsaitis2015cross,baltruvsaitis2013dimensional} & \checkmark &  & \checkmark & & \checkmark & \checkmark & \checkmark & \checkmark & \checkmark \\
    \midrule
    OpenFace 2.0~\cite{baltrusaitis2018openface} & ~\cite{wood2015rendering,zadeh2017convolutional,zhang2016joint} & \checkmark &  & \checkmark & & \checkmark & \checkmark & \checkmark & \checkmark & \checkmark \\
    \midrule
    \textbf{LibreFace (Ours)} & ~\cite{he2016deep,he2022masked,yang2021knowledge} & \checkmark & \checkmark & \checkmark & \checkmark & \checkmark & \checkmark & \checkmark & \checkmark & \checkmark \\
    \bottomrule
    \end{tabular}}
    \vspace{-10px}
    \label{tab:toolboxes}
\end{table*}

\section{Related Work}
\subsection{Pre-trained Vision Encoders}
Previous studies have proposed various pre-trained vision encoders to extract strong representations for computer vision tasks, including CNN-based models, \eg, Residual Network (ResNet) \cite{he2016deep}, ResNeXt \cite{xie2017aggregated}, and transformer-based approaches, \eg, Vision Transformer (ViT) \cite{dosovitskiy2020image}, Swin Transformer \cite{liu2021swin} and Masked Auto-encoder (MAE) \cite{he2022masked}.

He \etal propose Residual Network (ResNet) \cite{he2016deep}, a CNN-based model, to solve the gradient vanishing problem. He \etal introduce the residual block in which there is a direct connection skipping some layers in between, \ie, skip connection. 
Swin Transformer \cite{liu2021swin} is a hierarchical vision transformer and can serve as a general-purpose backbone for computer vision. It has gained significant attention due to its strong performance in different recognition tasks. One of its key features is the use of shifted windows, which enables cross-window connections and leads to more efficient self-attention computation. Moreover, the hierarchical structure allows it to handle facial features of various scales. 
Another noteworthy work is the self-supervised Masked Auto-encoder (MAE) \cite{he2022masked}. To eliminate the limitation of lacking labeled training data, the training process is conducted through an image reconstruction task. Subsets of image patches are masked randomly, and the remaining patches are passed to the encoder. A small decoder then processes the encoded patches and masked tokens to reconstruct the original image. After pre-training, the decoder is discarded, and the encoder can be applied to downstream classification tasks. In this paper, we extract the features with ResNet, Swin Transformer, and MAE for facial expression analysis. All of them are pre-trained on ImageNet \cite{deng2009imagenet}, which is not a face dataset.

Recently, Cai \etal propose MARLIN \cite{cai2023marlin}, an extension of MAE for facial video representation learning. MARLIN is pre-trained on the YouTube Face dataset \cite{wolf2011face} with video reconstruction, and it learns universal features which are transferable across various facial analysis tasks. 
Thus, motivated by MARLIN, we argue that further pre-training the vision encoders on various face datasets, \eg, EmotioNet \cite{fabian2016emotionet}, AffectNet \cite{kollias2021affect}, and FFHQ \cite{karras2019style} can boost the performance for facial expression analysis.

\subsection{Facial Expression Analysis Toolkit}
Over the past few years, several open-source toolkits for facial behavior analysis have been developed with traditional machine learning methods, \eg, histogram of gradients (HOG) and support vector machine (SVM). The most popular tools among these are OpenFace \cite{baltruvsaitis2016openface} and OpenFace 2.0 \cite{baltrusaitis2018openface}.

OpenFace \cite{baltruvsaitis2016openface} is an open-source framework for facial behavior analysis. The framework is capable of various facial analysis tasks such as landmark detection, gaze tracking, and AU detection.
OpenFace 2.0 \cite{baltrusaitis2018openface} is an extension work of OpenFace with several improvements over the original one. The capabilities of OpenFace 2.0 include facial landmark detection, head pose estimation, eye-gaze estimation, and facial action unit recognition.

With this work, we aim to provide a more accurate and efficient facial expression analysis framework based on deep neural networks, \eg, ResNet \cite{he2016deep}, Swin Transformer \cite{liu2021swin} and MAE \cite{he2022masked}, which have recently achieved resounding success in other computer vision tasks. We also deploy the idea of traditional knowledge distillation \cite{hinton2015distilling} and utilize feature-wise distillation based on \cite{zhang2021improve, yang2021knowledge} to further improve the inference efficiency to achieve real-time facial expression analysis better than the previous works.

\section{Problem Formulation}
\subsection{AU Intensity Estimation}
AU intensity estimation refers to the process of quantifying the intensity or strength of specific facial muscle movements known as Action Units (AUs). It involves analyzing facial images or video frames to detect and track the presence and magnitude of specific AUs. The intensity estimation is usually represented on a scale of from 0 to 5, with the bigger value indicating stronger strength of the AU activation. 
The approaches for AU intensity estimation can be categorized into traditional feature-based methods and deep learning-based methods. Traditional feature-based methods involve extracting hand-crafted features from facial images, such as texture, shape, or motion information. These features are then used as input to machine learning algorithms, such as support vector machines (SVMs) or random forests, to estimate AU intensities. OpenFace~\cite{baltruvsaitis2016openface} and OpenFace 2.0~\cite{baltrusaitis2018openface} are representative works of feature-based methods.

Deep learning-based methods leverage the power of deep neural networks to learn features from facial images or video frames automatically. Convolutional Neural Networks (CNNs), Recurrent Neural Networks (RNNs), and Transformers structures can be used for this task. Affdex 2.0~\cite{bishay2022affdex} is a representative work of this category. In LibreFace, we use Masked Auto-encoder (MAE) and ResNet-18 for AU intensity estimation. In Table ~\ref{tab:aus}, we provide the AUs that LibreFace can predict.

\begin{table}[t]
    \small
    \centering
    \caption{Available action units in LibreFace. \textbf{R} - LibreFace provides intensity estimation, \textbf{D} - LibreFace provides detection of occurrence.}
    \scalebox{1}{\begin{tabular}{l|l|c}
    \toprule
    \rowcolor{Gray}   
    AU & FACS name & Prediction \\
    \midrule
    AU1 & Inner brow raiser & R\\
    AU2 & Outer brow raiser& R\\
    AU4 & Brow lowerer& R\\
    AU5 & Upper lid raiser & R\\
    AU6 & Cheek raiser& R\\
    AU7 & Lid tightener & D\\
    AU9 & Nose wrinkler & R\\
    AU10 & Upper lip raiser & D\\
    AU12 & Lip corner puller & R\\
    AU14 & Dimpler & D\\
    AU15 & Lip corner depressor & R\\
    AU17 & Chin raiser & R\\
    AU20 & Lip stretcher & R\\
    AU23 & Lip tightener & D\\
    AU24 & Lip pressor & D\\
    AU25 & Lips part & R\\
    AU26 & Jaw drop & R\\
    \bottomrule
    \end{tabular}}
    \vspace{-20px}
    \label{tab:aus}
\end{table}

\subsection{Facial Expression Recognition}
\begin{figure}[t]
\centering
  \includegraphics[width=0.9\linewidth]{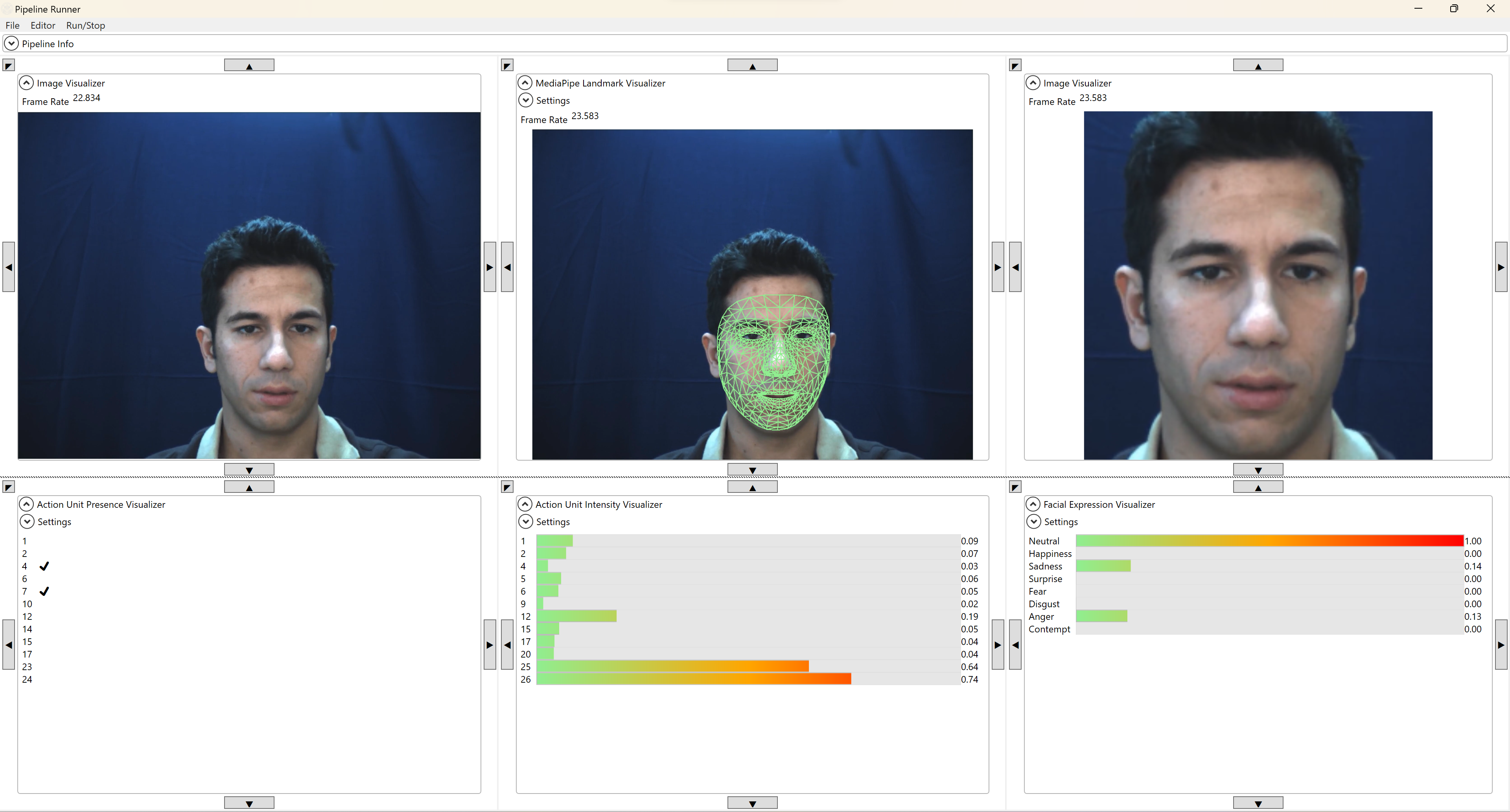}
  \caption{Overview of LibreFace software deployed in the Windows Operating System. We offer real-time facial action unit intensity estimation, detection, and facial expression recognition. }
  \label{fig:software}
  \vspace{-10pt}
\end{figure}

Facial Expression Recognition (FER) is the process of categorizing or recognizing the prototypical facial displays or expressions. It involves analyzing facial features, such as the position and movement of facial muscles, to identify and label specific emotions or expressions. It is worth noting that FER is a complex task due to factors like inter-individual variability, context dependency, and cultural differences in expressions. Researchers continue to explore advanced techniques~\cite{zhao2017deeply,yu2018generative,li2018occlusion}, such as deep learning architectures (\eg, CNNs, RNNs and Transformers), multi-modal approaches (combining facial and audio/text cues), and transfer learning, to improve the accuracy and robustness of FER systems. In Figure~\ref{fig:software}, we visualize the software of the LibreFace system deployed on a Personal Computer and show the example of feeding a local image file as input.

\section{Method}
\subsection{Overview}
The overall framework is illustrated in Figure~\ref{fig:pipeline}. We first discuss the image pre-processing, which involves face mesh and landmark detection and image alignment in Section~\ref{landmark}. Next, we feed the pre-processed images into a pre-trained MAE encoder, followed by a linear regression or classification layer that predicts the AU intensity values or facial expression labels, as detailed in Section~\ref{pre-train}. Once the MAE is fine-tuned, we employ feature-wise distillation to transfer the teacher model's (MAE) knowledge to a lightweight student model (ResNet-18), as outlined in Section~\ref{distillation}. Finally, we use the distilled ResNet-18 for efficient AU intensity estimation and FER.

\begin{figure*}
     \centering
     \begin{subfigure}[b]{0.45\textwidth}
         \centering
         \includegraphics[width=0.9\textwidth]{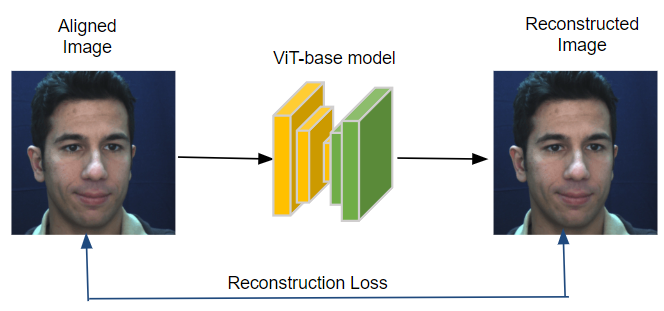}
         \caption{The ViT-base backbone of the MAE is pre-trained on the EmotionNet~\cite{fabian2016emotionet} dataset with a self-supervised image reconstruction loss.}
         \label{fig:EmotioNet-pretrain}
     \end{subfigure}
     \hfill
     \begin{subfigure}[b]{0.45\textwidth}
         \centering
        \includegraphics[width=0.9\linewidth]{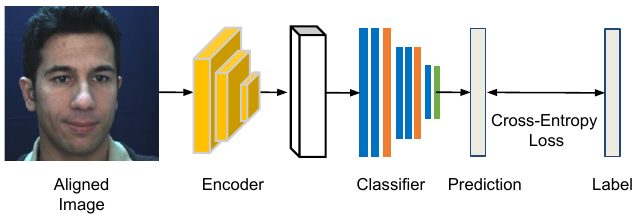}
         \caption{We pre-train the whole pipeline with MAE or ResNet-18 encoder on the AffectNet~\cite{mollahosseini2017affectnet} and FFHQ~\cite{karras2019style} dataset with cross-entropy loss.}
         \label{fig:ffhq-pertrain}
     \end{subfigure}
        \caption{Pre-training strategies of LibreFace for facial expression analysis.}
        \label{fig:pre-train}
        \vspace{-10pt}
\end{figure*}

\begin{figure}[t]
\centering
  \includegraphics[width=0.9\linewidth]{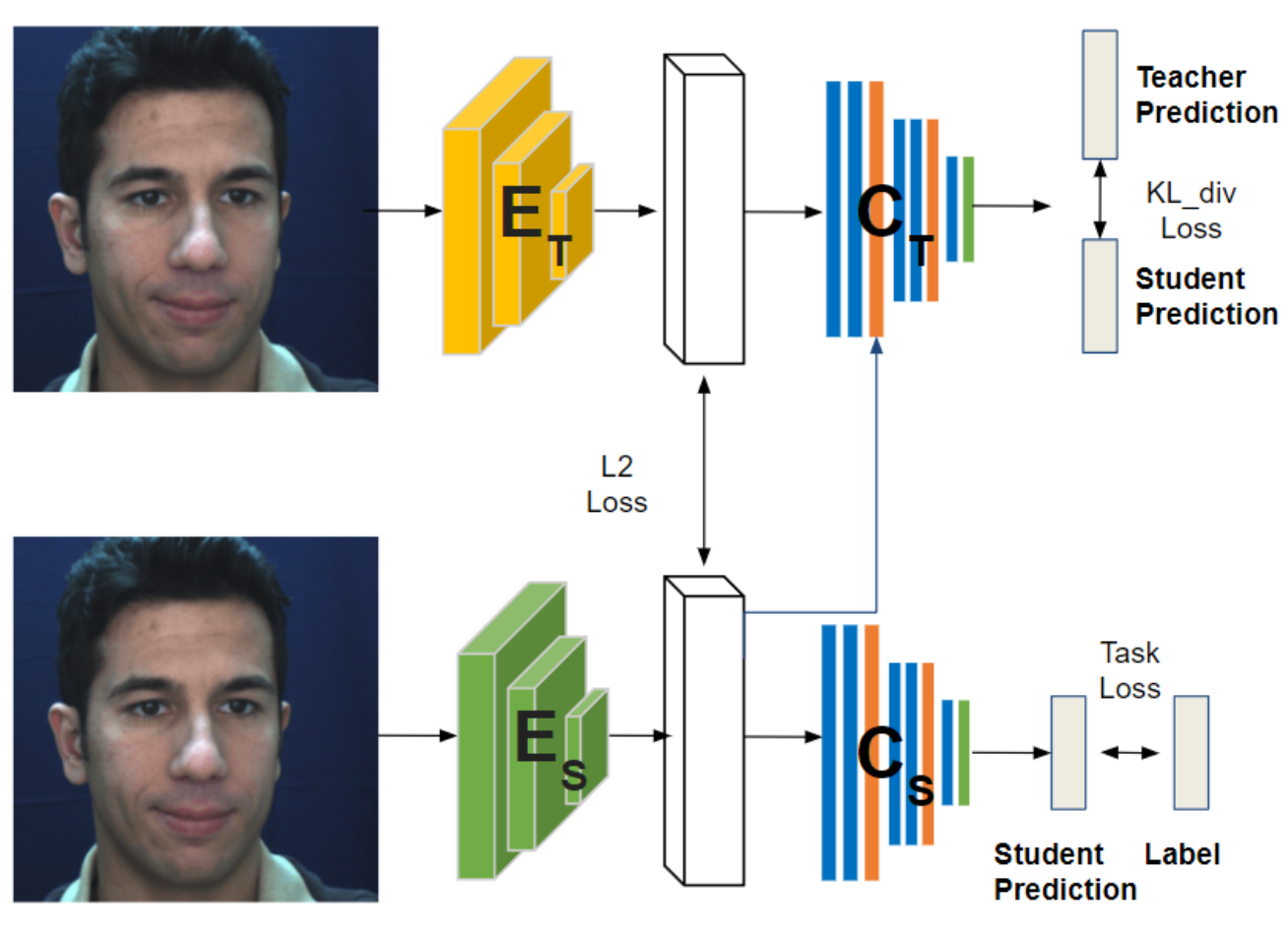}
  \caption{Overview of feature-wise distillation. We train the student model encoder $E_{S}$ with feature-wise distillation, which transfers the knowledge from the teacher model encoder $E_{T}$. $C_{T}$ and $C_{S}$ represent the classifier of the teacher and student model, respectively.}
  \label{fig:distillation}
  \vspace{-10pt}
\end{figure}

\subsection{Face Mesh and Landmark Detection}
\label{landmark}
Prior to inputting an arbitrary image into deep neural networks, it is imperative to perform facial image alignment based on localized landmarks. Facial image alignment involves geometric transformations, such as translation, rotation, and scaling, to convert the input face image into a canonical or standardized form. This process ensures consistent positioning of facial features across various images, facilitating the learning of patterns by our model. In our methodology, we leverage an open-source tool, \ie, MediaPipe~\cite{48292}, to detect the face mesh within the input image. By utilizing the detected mesh, we determine the precise locations of the facial landmarks, including the eyes, eyebrows, nose, and mouth. Subsequently, these facial landmarks are utilized to crop and align the facial region, culminating in the production of the resulting aligned output.

\subsection{Pre-training and Fine-tuning}
\label{pre-train}
Pre-training in facial expression analysis tasks refers to the process of training a deep learning model on a large dataset, typically using a supervised or self-supervised learning approach, before fine-tuning it on the specific tasks. The rationale behind pre-training is to leverage the general knowledge learned from a large-scale dataset to improve the performance and efficiency of the model on the target task, which allows the model to learn general visual representations, overcome data scarcity, facilitate transfer learning, improve feature extraction, and enhance the robustness and generalization capabilities. 

In this section, we discuss the pre-training process for the MAE teacher model and the ResNet-18 student model. As illustrated in Figure \ref{fig:pre-train}, the ViT-base structural Masked Auto-encoder (MAE) is pre-trained on EmotioNet \cite{fabian2016emotionet}. In this work, we use the model weights provided by \cite{zhang2022emotion}. Subsequently, we employ the weights of the ViT encoder as the backbone, add a linear classifier, and further pre-train the entire model on AffectNet\cite{kollias2021affect} and FFHQ~\cite{karras2019style} datasets. After pre-training on these two datasets, we fine-tune the model on DISFA~\cite{mavadati2013disfa} for AU intensity estimation. We use a Mean Squared Error (MSE) loss as the task-specific loss for AU intensity estimation since it is a regression task, which can be expressed as 
\begin{equation}
\label{eq:1}
\mathcal{L}_{Task} = \parallel \hat{y} - y\parallel^{2},
\end{equation} 
where $\hat{y}$ and $y$ represent the prediction and ground-truth for AU intensities.

Similarly, we employ ResNet-18 as the encoder and train it on AffectNet~\cite{kollias2021affect} and FFHQ~\cite{karras2019style} along with a linear classifier. Subsequently, we fine-tune the whole model on DISFA~\cite{mavadati2013disfa}. More details can be found in Section~\ref{expsetup}. We use a Cross-Entropy (CE) loss for the classification of FER. The task-specific loss is defined as
\begin{equation}
\label{eq:2}
\mathcal{L}_{Task} = -\sum_{c=1}^My\log(\hat{y}),
\end{equation}
where $M$ is the number of categories of facial expressions, while $\hat{y}$ and $y$ represent one-hot encoded prediction and ground-truth labels.

\subsection{Feature-wise Distillation}
\label{distillation}
Given that MAE is constructed on a ViT-base backbone, which demands significant computational resources, we aim to enhance the pipeline's efficiency and speed during inference. Although ResNet-18 already demonstrates strong performance for AU intensity estimation, we can further distill knowledge from the MAE into ResNet-18 to simultaneously boost performance and reduce computational cost. As illustrated in Figure~\ref{fig:distillation}, we propose to distill in a feature-matching manner, following the idea in~\cite{yang2021knowledge}. During distillation, all the weights from the teacher model, including an encoder $\text{E}_{T}$ and a classifier $\text{C}_{T}$, are frozen, and we fine-tune the student model encoder $\text{E}_{S}$ and classifier $\text{C}_{s}$ with a feature-matching loss $\mathcal{L}_{FM}$, a task-specific regression or classification loss $\mathcal{L}_{Task}$ (Equations \ref{eq:1} and \ref{eq:2}) and a Kullback–Leibler (KL) divergence loss $L_{KL}$.

We minimize the difference between the features from teacher encoder $f_{T}$ and the features from student encoder $f_{S}$ by a simple MSE loss
\begin{equation}
\mathcal{L}_{FM} = \parallel f_{T} - \text{I}(f_{S}) \parallel ^{2},
\end{equation}
\noindent where $\text{I}(\cdot)$ denotes the linear interpolation function to match the feature dimension.

$\mathcal{L}_{Task}$ is the same as any model trained in a non-distillation fashion (Equations \ref{eq:1} and \ref{eq:2}), which corrects the wrong prediction from $\textbf{C}_{S}$ with ground-truth labels. 

In the original knowledge distillation approach, as proposed by Hinton \etal ~\cite{hinton2015distilling}, the KL divergence loss minimizes the discrepancy between the teacher and student classifiers. In our work, however, we input the student features into the teacher classifier, utilizing the frozen, pre-trained teacher classifier for both the teacher and student models. This approach grants more flexibility to the optimization algorithm, allowing it to adjust the weights of both student encoder $\text{E}_{S}$ and classifier $\text{C}_{S}$ to minimize the loss more effectively.

\begin{equation}
L_{KL}=-\text{Softmax}\left(\hat{y}_{T}\right) \log \text{Softmax}\left(\hat{y}_{S}\right)
\end{equation}
where $\hat{y}_{T}$ and $\hat{y}_{S}$ denote the predictions from $\text{C}_{T}$ with teacher features and student features respectively.

Finally, we train the whole pipeline with the following overall loss
\begin{equation}
\mathcal{L} = \mathcal{L}_{FM} + \alpha \mathcal{L}_{Task} +\beta \mathcal{L}_{KL},
\end{equation}
\noindent where $\alpha$ and $\beta$ are the loss weights.

\section{Experiments}

\subsection{Dataset}

\begin{table*}[t]
    \small
    \centering
    \caption{Performance of our proposed LibreFace pipeline on DISFA~\cite{mavadati2013disfa}. Following~\cite{baltruvsaitis2016openface,baltrusaitis2018openface}, we evaluate our pipeline with five-fold cross-validation. We use Pearson Correlation Coefficient (PCC) for comparison to other methods.}
    \resizebox{\linewidth}{!}
    {\begin{tabular}{l|cccccccccccc|c}
    \toprule
    \rowcolor{Gray}
    Method & AU1 &AU2& AU4 &AU5& AU6 &AU9 &AU12 &AU15& AU17& AU20& AU25& AU26 & \textbf{Avg.} $\uparrow$ \\
    \midrule
    CNN~\cite{7284873} & 0.60 &0.53& 0.64 &0.38 &0.55 &0.59& 0.85& 0.22 &0.37 &0.15 &0.88 &0.60 &0.53 \\
    D-CNN~\cite{7780738} & 0.49& 0.39& 0.62& 0.44& 0.53& 0.55& 0.85& 0.25& 0.41& 0.19& 0.87& 0.59& 0.51 \\
    OpenFace 2.0~\cite{baltrusaitis2018openface} & \textbf{0.64} & 0.50 &0.70 &\textbf{0.67}& 0.59 &0.54& 0.85& \textbf{0.39 }&0.49& 0.22 &0.85 &0.67& 0.59 \\
    \textbf{LibreFace (Ours)} & 0.63 & \textbf{0.72}& \textbf{0.78} & 0.59 & \textbf{0.59}&\textbf{ 0.62}&\textbf{0.85}& 0.36 &\textbf{0.50} &\textbf{0.34}& \textbf{0.94}&\textbf{0.68} &\textbf{0.63}  \\
    \bottomrule
    \end{tabular}}
    \label{tab:DISFA}
    \vspace{-10pt}
\end{table*}

\begin{table*}[t]
    \small
    \centering
    \caption{Performance of our pipeline with different encoders and distillation on BP4D~\cite{zhang2014bp4d}. Res-18 is our baseline model without any pre-training or distillation. MAE\_ViT\_base denotes the encoder is replaced by MAE encoder with ViT-base structure being pre-trained. We report the cross-validated F1 score as the evaluation metric for all experiments. Shao \textit{et al.} \cite{shao2018deep} and Luo \textit{et al.} \cite{luo2022learning} are two sota methods that share the same evaluation setting. }
    \resizebox{\linewidth}{!}{\begin{tabular}{l|cccccccccccc|c}
    \toprule
    \rowcolor{Gray}
    Method &AU1 & AU2 & AU4& AU6& AU7& AU10& AU12& AU14& AU15& AU17& AU23& AU24 & \textbf{Avg.} $\uparrow$ \\
    \midrule
    Shao \textit{et al.} \cite{shao2018deep} & 47.2 & 44.0 & 54.9 & 77.5 & 74.6 & \textbf{84.0} & 86.9 & 61.9& 43.6 & 60.3 & 42.7 & 41.9 & 60.0 \\
    Luo \textit{et al.} \cite{luo2022learning} & 53.7& 46.9 & 59.0 &\textbf{ 78.5}& \textbf{80.0} & 84.4 & \textbf{87.8} & \textbf{67.3} & \textbf{52.5} & 63.2& 50.6 & 52.4 & \textbf{64.7}
\\
    Res-18 & 19.6 &13.9 &33.8 &75.8 &71.1 &80.5 &86.4 &50.3 &23.4 &59.0 &21.1 &31.2 & 47.2 \\
    MAE\_ViT\_base &\textbf{57.0} &\textbf{51.1} &\textbf{61.1}& 76.4 &76.1& 79.1 &86.3& 48.3 &\textbf{52.5} &\textbf{64.7}& \textbf{51.3} &\textbf{54.0 }& 63.2 \\

    \textbf{LibreFace (Ours)} &49.9& 47.8& 56.5 &77.9 &79.6& \textbf{84.0}&87.0& 59.0&46.0&63.1 &43.2& 49.9 & 62.0 \\ 
    \bottomrule
    \end{tabular}}
    \label{tab:BP4D}
    \vspace{-15pt}
\end{table*}

\begin{table}[t]
    \small
    \centering
    \caption{Performance of facial expression recognition leveraging our distillated and pre-trained ResNet-18 model evaluated on AffectNet~\cite{mollahosseini2017affectnet} and RAF-DB~\cite{li2017reliable} with overall accuracy and comparison to a few state-of-the-art methods. We report the performance of sota methods.} 
    \scalebox{1}{\begin{tabular}{l|ccc}
    \toprule
    \rowcolor{Gray}   
    Method & AffectNet$\uparrow$ & RAF-DB$\uparrow$ \\
    \midrule
    VGG-16~\cite{simonyan2014very} & 51.11 & 80.96\\
    DLP-CNN~\cite{zhao2017deeply} & 54.47 & 80.89\\
    GAN-Inpainting~\cite{yu2018generative} & 52.97 &81.87\\
    gaCNN~\cite{li2018occlusion} & \textbf{58.78} & \textbf{85.07} \\
    \midrule
    \textbf{LibreFace (Ours)} & 49.71 & 82.79 \\
    \bottomrule
    \end{tabular}}
    \label{tab:fer}
    \vspace{-15pt}
\end{table}

\begin{table}[t]
    \small
    \centering
    \caption{Ablation of our proposed pipeline with different pre-training and encoder settings on DISFA~\cite{mavadati2013disfa}. We also evaluate our pipeline with five-fold cross-validation and use Pearson Correlation Coefficient (PCC), Mean Absolute Error (MAE), and Mean Squared Error (MSE) as the evaluation metric for ablation analysis. \textbf{Pre-train} denotes the corresponding encoder is pre-trained on AffectNet~\cite{mollahosseini2017affectnet} and FFHQ~\cite{karras2019style}. \textbf{Distill} denotes the model is trained with the original distillation manner, and \textbf{FM\_distill} denotes the model is trained with feature-matching distillation. We also use Swin-Transformer tiny structure and Masked Auto-encoder with Vision-Transformer base structure as the encoder, represented as Swin-Tiny and MAE\_ViT\_base.  \textbf{Res-18+FM\_distill} is the final model we deployed in LibreFace.  }
    \scalebox{1}{\begin{tabular}{l|ccc}
    \toprule
    \rowcolor{Gray}
    Method & PCC$\uparrow$ & MAE$\downarrow$ & MSE$\downarrow$\\
    \midrule
    Res-18 & 0.518 & 0.278 & 0.352 \\
    Res-18+Pre-train & 0.614 &\textbf{ 0.236}& 0.260 \\
    Res-18+Distill & 0.620 & 0.896 & 0.550 \\
    Res-18+FM\_distill & \textbf{0.628} &0.244 & \textbf{0.260}  \\
        \midrule
    Swin-Tiny & 0.606 & 0.252 & 0.286  \\
    Swin-Tiny+Pre-train & \textbf{0.644} & \textbf{0.236} & \textbf{0.246} \\
        \midrule
    MAE\_ViT\_base & 0.668 & \textbf{0.202} & 0.272  \\
    MAE\_ViT\_base+Pre-train & \textbf{0.674} & \textbf{0.202} & \textbf{0.270} \\
    \bottomrule
    \end{tabular}}
    \label{tab:ablation}
    \vspace{-5pt}
\end{table}

\textbf{EmotionNet~\cite{fabian2016emotionet}}, \textbf{FFHQ~\cite{karras2019style}} and \textbf{AffectNet~\cite{mollahosseini2017affectnet}} are used to pre-train our models.  EmotionNet dataset~\cite{fabian2016emotionet} is a large-scale collection of facial expressions in the wild, containing approximately 975,000 annotated images. The annotations include Action Units (AUs) and 421 emotion keywords, which are automatically assigned using the EmotionNet algorithm~\cite{fabian2016emotionet}. This dataset is suitable for a variety of AU applications, including recognition and intensity estimation, as well as basic and compound emotion recognition. Mixed results may occur when facial expressions are not well composed or ambiguous. For instance, some annotations may indicate a combination of multiple emotions or an uncertain degree of intensity, such as 68.66\% disgust and 27.98\% fear.

AffectNet~\cite{fabian2016emotionet} is a vast facial expression dataset comprising over 1,000,000 facial images. The images are gathered from the internet by searching for 1250 keywords related to emotions in six different languages. To support both discrete emotion classification and continuous affective computing, the dataset is manually labeled for eight different facial expressions and the intensity of valence and arousal. It is currently the largest database of facial expression, valence, and arousal, providing a valuable resource for research in affective computing. FFHQ~\cite{karras2019style} is a dataset of high-quality human facial images. Its initial purpose was to serve as a benchmark for the evaluation of generative adversarial networks (GANs). This dataset provides 70,000 images in the PNG format with a resolution of 1024×1024 and a substantial range of variation with respect to age, ethnicity, and image background. Furthermore, FFHQ boasts extensive coverage of facial accessories such as eyeglasses, sunglasses, and hats. 

\textbf{DISFA~\cite{mavadati2013disfa}} is utilized for training and evaluation of AU intensity estimation. This dataset is a non-posed large-scale facial expression database captured by stereo videos of 27 adult subjects (12 females and 15 males) from different ethnicities. These videos are annotated with facial action unit intensity values ranging between $[0, 5]$ manually labeled by two human FACS experts. Following the setting from OpenFace 2.0~\cite{baltrusaitis2018openface}, we train our proposed pipeline with five-fold cross-validation so that we fairly compare our results.  We apply the same model we used for AU intensity estimation to \textbf{BP4D}~\cite{zhang2014bp4d} dataset to provide AU detection results for those AUs that are not provided with labels in DISFA~\cite{mavadati2013disfa}.

\begin{table}[t]
    \small
    \centering
    \caption{Efficiency Comparison between LibreFace and OpenFace 2.0~\cite{baltrusaitis2018openface}. We report the average (Avg.) and standard deviation (Std.) of durations of each round of tests and averaged frame rate per second (FPS).}
    \scalebox{1}{\begin{tabular}{l|cccc}
    \toprule
    \rowcolor{Gray}   
    Method & Avg.$\downarrow$ & Std. & FPS$\uparrow$\\
    \midrule
    OpenFace 2.0~\cite{baltrusaitis2018openface} (AU only) & 50.11 & 0.22 & 19.96 \\
    \textbf{LibreFace (AU only)} & \textbf{25.11} & 0.60 & \textbf{39.82} \\
    \textbf{LibreFace (Ours)} & 37.20 & 0.66 & 26.88 \\
    \bottomrule
    \end{tabular}}
    \label{tab:efficiency_compare}
    \vspace{-10pt}
\end{table}

\begin{table}[t]
    \small
    \centering
    \caption{Efficiency and exported executable model size comparison among different encoders in our proposed pipeline. We report the average (Avg.) and standard deviation (Std.) of durations of each round of test, averaged frame rate per second (FPS), and model size in MB(Size).}
    \scalebox{1}{\begin{tabular}{l|ccccc}
    \toprule
    \rowcolor{Gray}   
    Method & Avg.$\downarrow$ & Std. & FPS$\uparrow$ & Size$\downarrow$ \\
    \midrule
    Res-18 & \textbf{37.20} & 0.66 & \textbf{26.88 } & \textbf{43}\\
    Swin-Tiny & 55.94 & 1.45 & 17.88 & 185 \\
     MAE\_ViT\_base & 87.93 & 3.96 & 11.37 & 403\\
    \bottomrule
    \end{tabular}}
    \label{tab:efficiency_ablation}
    \vspace{-10pt}
\end{table}

\begin{table}[t]
    \small
    \centering
    \caption{Efficiency comparison among CPU-only mode and GPU mode in our proposed pipeline. We report the average (Avg.) and standard deviation (Std.) of durations of each round of test, and averaged frame rate per second (FPS)}
    \scalebox{1}{\begin{tabular}{l|ccccc}
    \toprule
    \rowcolor{Gray}   
    Method & Avg.$\downarrow$ & Std. & FPS$\uparrow$ \\
    \midrule
    CPU-only & 37.20 & 0.66 & 26.88  \\
    GPU & \textbf{6.07} & 0.13 & \textbf{164.82 } \\
    \bottomrule
    \end{tabular}}
    \label{tab:efficiency_compare_processor}
    \vspace{-10pt}
\end{table}

For FER, we report the result of our proposed LibreFace on the test set of 
AffectNet~\cite{mollahosseini2017affectnet} and \textbf{RAF-DB~\cite{li2017reliable}}. We apply the same pre-training strategy and feature-wise distillation we used for AU intensity estimation to FER.

RAF-DB~\cite{li2017reliable} is a large-scale facial expression dataset consisting of approximately 30,000 diverse facial images which feature a wide range of variations in age, gender, ethnicity, head pose, and lighting conditions. This work uses the single-label subset, including seven classes of basic emotions, for evaluation purposes.

\subsection{Implementation and Training Details}
All methods are implemented in PyTorch \cite{paszke17}. Training code and model weights are available for the sake of reproducibility. All tasks are completed with one NVIDIA RTX 8000 GPU for training, and our pipeline can be tested with either a CPU-only machine or a machine with a GPU. 

The input image is first resized to $256\times256$. For data augmentation, each face is randomly cropped into $224\times224$ and randomly flipped in the horizontal direction. 
We train the model with the AdamW optimizer \cite{loshchilov2017decoupled} with a batch size of 128. The initial learning rate is 3e-5 with a weight decay of 1e-4. We train the model for a maximum of 20 epochs with early stopping. The model is trained with MSE loss for AU intensity estimation and cross-entropy loss for FER while being evaluated on the validation set at the end of every epoch. For the loss weights in the feature-wise distillation, we set $\alpha=1.0$ and $\beta = 1.0$ for all experiments after hyper-parameter tuning. 

\subsection{Software Implementation}
\label{software}
The software is implemented in C\# and consists of three major components, MediaPipe \cite{48292}, image aligner, and ONNX models. MediaPipe is used for detecting faces and providing facial landmarks. The image aligner crops and aligns faces from input images according to detected facial landmarks. Aligned images will be provided as inputs for our models. Models run concurrently and can be enabled or disabled according to use cases. All components are developed as .NET libraries and are multi-platform. For easy deployment, we have also developed a graphical user interface and turned them into Microsoft's Platform for Situated Intelligence components \cite{psi} that can be easily used in OpenSense~\cite{opensense}.
\subsection{Experimental Setup}
\label{expsetup}
\noindent \textbf{Pre-training and Fine-tuning Strategy.} For the final model deployed in the LibreFace pipeline, we apply knowledge distillation and pre-training to boost performance. We first pre-train a Masked Auto-encoder on EmotioNet, AffectNet, and FFHQ training set. The training strategy is shown in Figure~\ref{fig:pre-train}. We then take the checkpoint of the MAE encoder and fine-tune it on DISFA~\cite{mavadati2013disfa} for AU intensity estimation. After we have finished the five-fold cross-validation for MAE, we adopt the MAE checkpoint (including the encoder and linear classifier) from these five folds as the teacher model and apply feature-wise knowledge distillation to train a ResNet-18 encoder and a linear classifier. For AU detection on BP4D~\cite{zhang2014bp4d}, we repeat the above pre-training and fine-tuning process and evaluate with three-fold cross-validation as in ~\cite{shao2018deep,luo2022learning}.

We apply the same pre-training strategy for MAE on FER. We take the checkpoint of both the encoder and linear classifier from the MAE trained on the training set of AffectNet and FFHQ, then apply feature-wise knowledge distillation to the ResNet-18 encoder and classifier with the previous MAE checkpoint as the teacher model. We train the pipeline on each training set of AffectNet, FFHQ, and RAF-DB, and report our performance on the test set of each dataset, respectively.

\subsection{Results}
\noindent
\textbf{AU Intensity Estimation.} We show the AU intensity estimation results of the proposed LibreFace on DISFA~\cite{mavadati2013disfa} in Table~\ref{tab:DISFA}. Since OpenFace 2.0~\cite{baltrusaitis2018openface} evaluates their model with five-fold cross-validation and reports Pearson Correlation Coefficient (PCC) of the results, we follow their setting to compare fairly. We also compare LibreFace to other methods following this setting.  LibreFace achieves 0.63 average PCC score across all AUs, while  OpenFace 2.0~\cite{baltrusaitis2018openface} achieves an average PCC of 0.59. In addition to achieving the best average PCC, our method also performs better than other methods on 9 out of total 12 AUs included in DISFA~\cite{mavadati2013disfa}.
Since the amount of labeled AUs is limited in DISFA~\cite{mavadati2013disfa}, we also fine-tune our model with different encoders on BP4D~\cite{zhang2014bp4d}, as shown in Table~\ref{tab:BP4D}, in order to provide AU detection (presence or absence of AUs) result for AUs not labeled in DISFA~\cite{mavadati2013disfa}. 

\noindent \textbf{Facial Expression Recognition.} We show the FER results in Table~\ref{tab:fer}, where overall accuracy ($\uparrow$) is the evaluation metric. We compare to other state-of-the-art methods according to their reported performance in the paper. 
We achieve comparable results to the state-of-the-art methods, which require far more computation and training/inference time than LibreFace.

\noindent \textbf{Ablation Study and Efficiency Analysis.} We provide ablation analysis when different encoders are implemented in our pipeline and their corresponding efficiency analysis. We also compare the computation cost of LibreFace to OpenFace 2.0~\cite{baltrusaitis2018openface} in this section. For the efficiency comparison, we run LibreFace with different encoders and OpenFace 2.0~\cite{baltrusaitis2018openface}. We run all experiments with \textbf{Intel i9-13900K} CPU, \textbf{64GB DDR5 4800MHz} RAM, and \textbf{Nvidia GTX1080} on \textbf{Windows 11} Operating System. E-cores and hyper-threading are disabled. We randomly pick 1000 images in a resolution of $1024 \times 768$ from our data and feed them to these toolboxes. We conduct five rounds of tests for each toolbox and calculate averaged running time (Avg.) in second and standard deviation (Std.). We also report the FPS and exported binary model size.

As shown in Table~\ref{tab:ablation}, utilizing a more complicated and heavier structure as an encoder, \eg, Swin-Transformer and Masked Auto-encoder, yields better performance. However, considering the efficiency factor shown in Table~\ref{tab:efficiency_ablation}, we finally adopt  ResNet-18 as the encoder and apply feature-wise knowledge distillation to boost the performance. The original knowledge distillation yields better PCC than our baseline (Res-18), but MAE and MSE are significantly worse. In contrast, feature-wise knowledge distillation apparently boosts the performance for all evaluation metrics.

From Table~\ref{tab:efficiency_compare}, we observe the efficiency improvement of LibreFace compared to OpenFace 2.0~\cite{baltrusaitis2018openface}. LibreFace provides more accurate facial expression analysis while running \textbf{two times} faster. Noted that \textit{AU only} stands for we only turn on the function of AU intensity estimation and AU Detection, with other output features turned off so that the efficiencies are comparable. We also report the efficiency performance of LibreFace while predicting AU, FER, and other outputs at the same time, where LibreFace still runs \textbf{1.5 times} faster than \cite{baltrusaitis2018openface}. In Table~\ref{tab:efficiency_compare_processor}, we observed that running our pipeline on GPUs is generally much faster than running on CPUs, which is in accordance with our expectations.

\section{Conclusions and Future work}

In this work, we introduce LibreFace, an accurate and flexible framework for facial expression analysis. With proposed feature-wise knowledge distillation and extensive pre-training, we offer a user-friendly toolbox for facial expression analysis. We leverage large-scale networks pre-trained on relevant tasks, then distillate the model to a lightweight structure with feature-wise knowledge to improve efficiency. We offer an easy-to-use toolbox better than OpenFace 2.0~\cite{baltrusaitis2018openface} in terms of wider applications, more accurate predictions, and comparable efficient inference.

The current build of our pipeline is tested on Windows platform. In the future, we plan to test our .NET assemblies on Linux and MacOS. All model checkpoints, including ResNet-18 after distillation, pre-trained Swin-Tiny, and pre-trained MAE, together with their corresponding training/inference code and the binary executable program are available on GitHub (link will be provided upon acceptance). As mentioned in the paper, LibreFace will be able to use in a CPU-only environment. We will further provide documents for LibreFace when the user is equipped with a GPU so that more efficient and faster inference will be accomplished with detailed installation and utilization instructions.

\section{Acknowledgement}
This work is sponsored by the U.S. Army Research Laboratory (ARL) under contract number W911NF-14-D-0005. The content of the information does not necessarily reflect the position or the policy of the Government, and no official endorsement should be inferred.
\clearpage
\newpage
{\small
\bibliographystyle{ieee_fullname}
\bibliography{egbib}
}

\end{document}


\title{Supplementary Materials for LibreFace}  

\maketitle
\thispagestyle{empty}
\appendix

\section{More Efficiency Analysis.}

{\small
\bibliographystyle{ieee_fullname}
\bibliography{egbib}
}